\documentclass[letterpaper, 10pt, conference]{ieeeconf}

\IEEEoverridecommandlockouts
\usepackage{amsmath, amssymb}
\usepackage{graphicx}
\graphicspath{{images/}}
\usepackage[adjust]{cite}
\usepackage{pgfplots}
\pgfplotsset{compat=1.8}
\usepackage{tikzscale}
\usepackage{xcolor}
\usepackage{xspace}
\usepackage{csquotes}
\usepackage{placeins}
\usepackage{algorithm}
\usepackage[noend]{algpseudocode}
\usepackage{mathabx}
\usepackage{siunitx}
\DeclareSIUnit{\fps}{fps}
\usepackage{paralist}
\usepackage[hidelinks]{hyperref}
\usepackage{subcaption}

\newcommand{\qwend}{\texttt{Qwen-2.5}\xspace}
\newcommand{\qwenu}{\texttt{Qwen-3}\xspace}
\newcommand{\botsort}{\texttt{BoT-SORT}\xspace}
\newcommand{\yolo}{\texttt{YOLOv8}\xspace}
\newcommand{\desc}{\texttt{TD-SP}\xspace}
\newcommand{\emb}{\texttt{VE-SP}\xspace}

\title{\LARGE \bf
Language-Based Swarm Perception: 
Decentralized Person Re-Identification via Natural Language Descriptions}

\author{Miquel Kegeleirs$^{1}$, Lorenzo Garattoni$^{2}$, Gianpiero Francesca$^{2}$, and Mauro Birattari$^{1}$
\thanks{$^{1}$MK and MB are with IRIDIA, Université libre de Bruxelles, Belgium. $^{2}$LG 
and GF are with Toyota Motor Europe.
Correspondence to: \texttt{mauro.birattari@ulb.be}}%
\thanks{The research received funding from Belgium’s Wallonia-Brussels Federation through the ARC Advanced Project \emph{GbO}. MB acknowledges support from the Belgian \emph{Fonds de la Recherche Scientifique}--FNRS.} 
\thanks{The idea was conceived by MK, LG, GF, and MB.
The experiments were designed by MK, LG, GF, and MB, and conducted by MK.
The original software was developed by MK.
The paper was drafted by MK and edited by MB.
All authors reviewed the manuscript and provided feedback.
The research was directed by MB.}%
}
\begin{document}

\maketitle
\thispagestyle{empty}
\pagestyle{empty}

\begin{abstract}
We introduce a method for decentralized person re-identification in robot swarms that leverages natural language as the primary representational modality.
Unlike traditional approaches that rely on opaque visual embeddings—high-dimensional feature vectors extracted from images—the proposed method uses human-readable language to represent observations.
Each robot locally detects and describes individuals using a vision-language model (VLM), producing textual descriptions of appearance instead of feature vectors.
These descriptions are compared and clustered across the swarm without centralized coordination, allowing robots to collaboratively group observations of the same individual.
Each cluster is distilled into a representative description by a language model, providing an interpretable, concise summary of the swarm’s collective perception.
This approach enables natural-language querying, enhances transparency, and supports explainable swarm behavior.
Preliminary experiments demonstrate competitive performance in identity consistency and interpretability compared to embedding-based methods, despite current limitations in text similarity and computational load.
Ongoing work explores refined similarity metrics, semantic navigation, and the extension of language-based perception to environmental elements. This work prioritizes decentralized perception and communication, while active navigation remains an open direction for future study.
\end{abstract}

\section{INTRODUCTION}

\noindent Swarm perception refers to the ability of a robot swarm~\cite{Ben2005sab,Sah2005sab,DorBirBra2014SCHOLAR} to leverage the sensory inputs of individual robots to achieve a collective understanding of the environment. 
Due to their distributed nature, robot swarms can collectively gather, share, and update information about their surroundings in a scalable, flexible, and fault-tolerant manner~\cite{BraFerBirDor2013SI}.
This can be particularly advantageous in people (re-)identification and tracking scenarios, especially in environments with unknown structure where static methods, which rely on strategic sensor placement or predefined path plan, are not applicable.
While previous work has demonstrated decentralized visual people re-identification using visual embeddings~\cite{KegGhaGar-etal2025arxiv}, we explore here a more interpretable and user-friendly alternative: using natural language descriptions.

The key idea is to equip each robot with the ability to generate textual descriptions of the people it observes---e.g., “a person wearing a black T-shirt and blue jeans”---using a vision-language model (VLM). 
These descriptions are then compared, allowing robots to identify and track individuals across space and time without exchanging raw visual data.

\smallskip\noindent
This approach provides two main benefits:
\begin{enumerate}
    \item It enables users to understand and inspect the knowledge held by the swarm using human language.
    \item It allows intuitive querying: instead of submitting a picture, a user can ask, e.g., “have you seen a person in a red hoodie?” and retrieve relevant images of that person.
\end{enumerate}
To manage redundancy, whenever a new textual description is generated, the robot checks whether a similar description of a person has been produced previously. If so, it determines whether the new description refers to the same individual or a different one. Similar descriptions are then merged, and a single representative summary is generated using a large language model (LLM).
This summary captures the shared attributes of the group in a concise and coherent natural language description.

This work aims to investigate whether natural language can serve as an effective medium for decentralized person re-identification in a robotic swarm.
Rather than optimizing for pursuit or coverage, we focus on evaluating the swarm’s capacity to form shared semantic representations under minimal coordination.
The goal is to lay the groundwork for future systems in which communication, reasoning, and user interaction can be conducted through interpretable language-based channels.

The proposed method is evaluated in the same simulated environment as previous work. 
While results are not yet optimal, they highlight the potential of natural language as a medium for decentralized swarm perception and communication.
This work focuses on the mechanisms of information collection and sharing; more advanced navigation strategies, such as following individuals or strategic repositioning, are considered out of scope at this stage and are left for future study.

\section{RELATED WORK}
\label{sec:related-work}

In swarm robotics, extensive research has focused on understanding collective behaviors~\cite{TriCam2015hci,BraFerBirDor2013SI,GarBir2016weeee} and collective decision-making~\cite{StrCasDor2018aamas,ValFerHam-etal2016AAMAS}, with perception consistently identified as a key enabler.
More broadly, collective perception has gained traction in domains such as (semi-)autonomous vehicles~\cite{GunTraWol2015itst,GunMenTra-etal2016vnc,ThaSepGoz2019ivs} and distributed monitoring systems~\cite{FioFehBod-etal2008JIRS,ChoSav2012eccv,MonAraLes-etal2023MTA}.
In these contexts, person re-identification~\cite{ZheTanHau2016CORR} plays a critical role, enabling agents to consistently recognize and track individuals across different views and locations to support data fusion.
Traditional re-identification pipelines rely heavily on deep learning~\cite{YeSheLin-etal2022}, particularly on feature embeddings trained using triplet loss and its variants~\cite{HerBeyLei2017arxiv}.
Although effective, these embeddings are inherently opaque, limiting transparency and making it difficult for users to interpret or audit the system’s internal representations.
Recent advances in vision-language models (VLMs) introduce the possibility of encoding visual information in natural language, offering interpretable and semantically rich alternatives to embedding vectors~\cite{RadKimHal-etal2021pmlr,LiZhaZha-etal2022cvpr,ZhaMusKol-etal2023iccv}.
Language-based interfaces have already enhanced human-robot interaction by allowing robots to follow high-level instructions and express their internal states in a more human-understandable form~\cite{DonZhaHua-etal2023arxiv,Atu2024arxiv,RahBahAbr-etal2025arxiv}.
While most re-ID research has historically focused on static CCTV surveillance~\cite{UkiMorHag2016CVIU,KoiMenCar-etal2017ecmr}, mobile robots offer advantages such as dynamic viewpoints, close-range sensing, and interactive capabilities~\cite{MurAts2018isis}.
These have been explored in single-robot systems for tasks such as face and voice-based re-identification~\cite{WanShePet-etal2019PRL,LuAshBer2024biorob}, as well as for person-following and user assistance~\cite{YeZhaPan-etal2023icra}.
Robust methods like CARPE-ID have demonstrated reliable person tracking despite occlusions and appearance changes~\cite{RolZunTsa-etal2024icra}.
More recently, multi-robot and swarm-based approaches to person re-identification have begun to emerge~\cite{PopGilMon-etal2022ir,KegGarLeg-etal2024icra,KegGarLeg-etal2024arxiv}, demonstrating promising scalability and resilience.

Our work builds on these developments by introducing a language-based approach to person re-identification in robot swarms.
Instead of opaque feature vectors, we employ  natural language descriptions, enabling interpretable cluster formation and intuitive querying. 
This approach aligns with growing interest in explainable AI and supports more transparent,  accessible interactions between humans and robotic systems~\cite{XuUszDu-etal2019nlpcc,DwiDavNai-etal2023ACMCS}.

\section{METHOD}
\label{sec:method}

In this work, we adopt a decentralized architecture that replaces opaque feature vectors with interpretable natural language descriptions.
Each robot independently detects, tracks, and describes individuals it observes using its onboard camera. 
Descriptions are clustered locally and refined through decentralized local communication with nearby peers.

\subsection{Local Data Acquisition}

Each robot processes its video stream using a three-stage pipeline (as illustrated in Figure~\ref{fig:acquisition}):
\begin{figure}[t]
    \centering
    \includegraphics[width=\columnwidth]{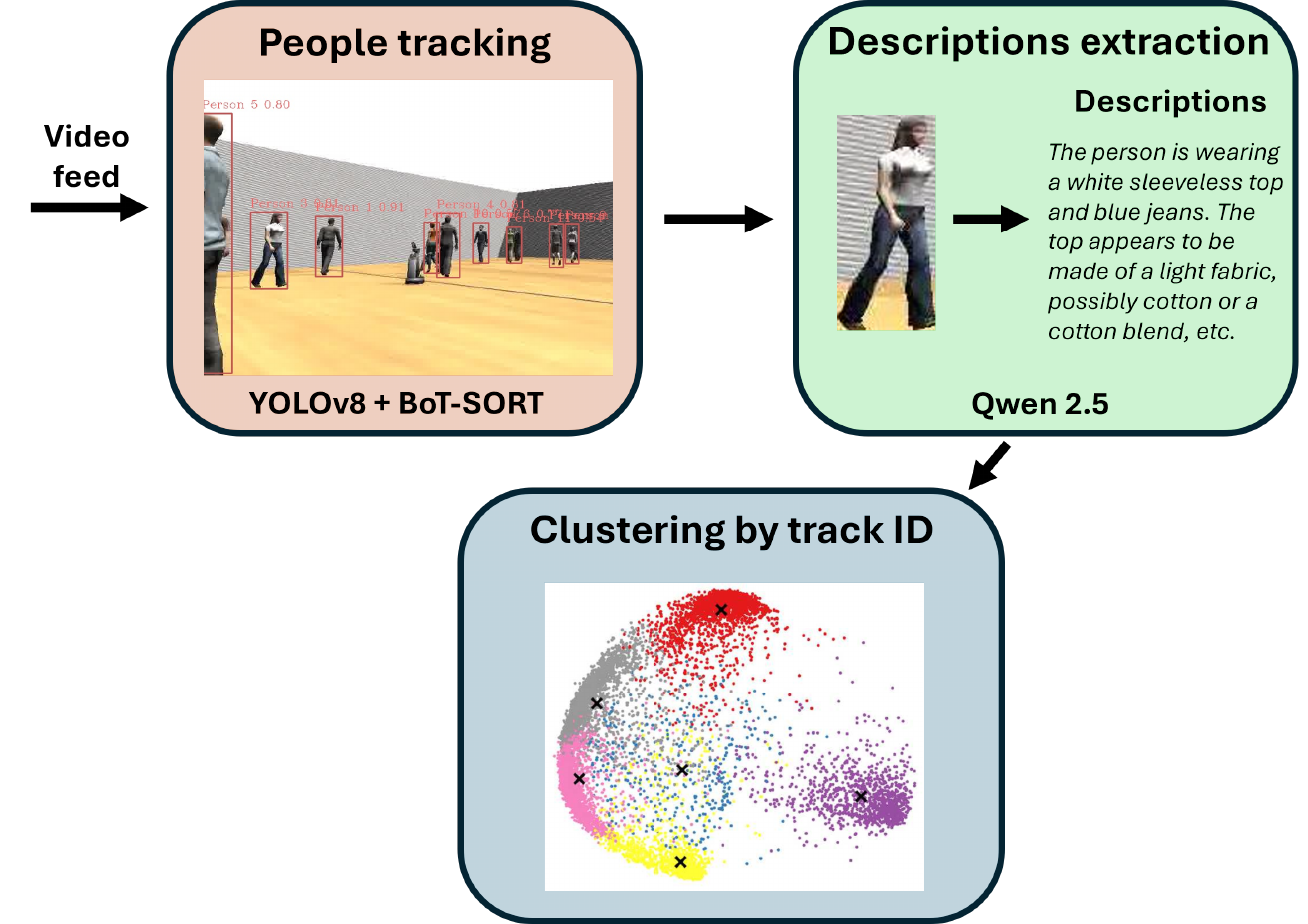}
    \caption{Data acquisition process.}
    \label{fig:acquisition}
\end{figure}
\begin{enumerate}
    \item \textbf{Detection:} People are detected using \yolo~\cite{VarSam2024adics}, trained on the COCO dataset~\cite{LinMaiBel-etal2014eccv}.
    \item \textbf{Tracking:} The \botsort algorithm~\cite{AhaOrfBob2022arxiv} assigns unique IDs to detected individuals and maintains tracks across frames.
    \item \textbf{Description Generation:} For each tracked person, a cropped image is passed to \qwend~\cite{qwen2.5}, which outputs a sentence-level natural language description (e.g., “a person in a blue shirt and gray pants”).
\end{enumerate}
In this work, two variants of the Qwen vision-language model are used:
\qwend{} is employed for generating descriptions of observed individuals from cropped images, while \qwenu{} is used to summarize multiple descriptions into a single, representative cluster-level description.
Although both are part of the Qwen family, they serve distinct roles in the pipeline.

\subsection{Description Similarity and Clustering}

Each robot maintains a local database of clusters, where each cluster represents a hypothesized individual and consists of one or more semantically similar descriptions.
Descriptions within a cluster are passed to \qwenu~\cite{BaiBaiChu-etal2023arxiv} to generate a concise, representative summary.
Whenever a new description is added to a cluster, the summary is updated accordingly via \qwenu.
Initially, descriptions are grouped according to tracking IDs provided by \botsort. 
Clusters associated with the same ID are merged as long as tracking remains consistent. 

When a new ID is to be assigned to a cluster, the corresponding description is compared to all existing clusters using cosine similarity computed on sentence-level embeddings.
If the similarity with an existing cluster exceeds a predefined threshold, the new description is merged into that cluster.
If no sufficiently similar cluster is found, a new cluster is created and initialized with the new description.

\subsection{Inter-Robot Communication and Merging}

\begin{figure}[t]
    \centering
    \includegraphics[width=0.9\columnwidth]{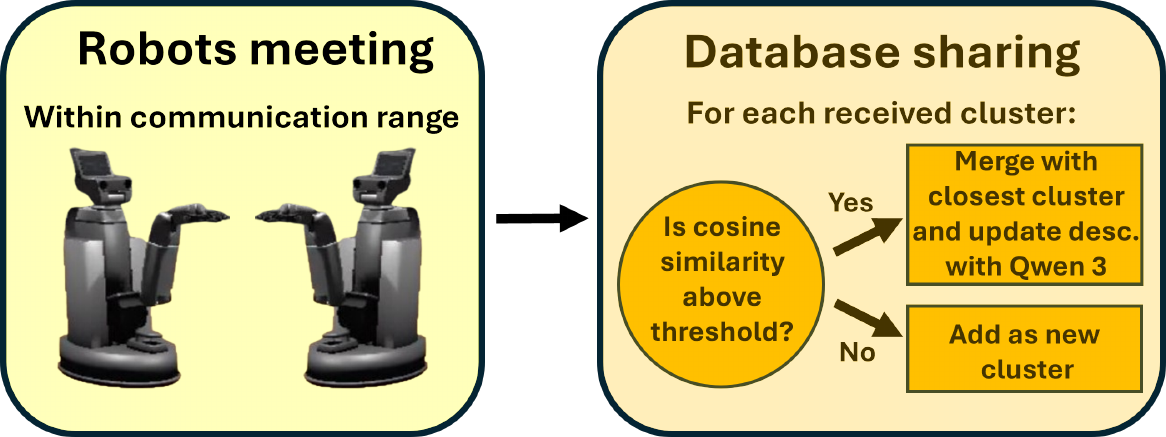}
    \caption{Data sharing process.}
    \label{fig:sharing}
\end{figure}

Robots periodically exchange their cluster databases upon encountering each other within communication range.
For each received cluster, a robot:
\begin{enumerate}
    \item compares its representative description to those of all local clusters using cosine similarity;
    \item if a match exceeds a fixed similarity threshold, the clusters are merged;
    \item otherwise, the received cluster is added as a new local entry.
\end{enumerate}
Following a merge, \qwenu is used to regenerate the representative description of the updated cluster.
Figure~\ref{fig:sharing} summarizes this process.
While our method only relies on textual descriptions for decentralized re-identification, robots may optionally share additional data---such as person images---depending on the application context. 
These images are not used for perception or clustering but can be useful for logging, visualization, or user-facing interfaces. 
Conversely, omitting image sharing may be preferable in privacy-sensitive scenarios.
In our case, images are shared solely to support the final demonstration (Section~\ref{sec:results}), where robots return visual samples from their clusters in response to natural-language queries. 
This sharing facilitates qualitative assessment but is not required for the algorithm to function.

\subsection{Exploration behavior}

In line with prior work~\cite{KegGhaGar-etal2025arxiv}, we adopt a random walk exploration strategy based on ballistic motion.
This provides a simple, decentralized, and unbiased method for spatial coverage. 
It serves as a neutral baseline that isolates the impact of decentralized communication and perception, avoiding confounding effects that would be introduced by more complex navigation behaviors.

\section{EXPERIMENTAL SETUP}
\label{sec:experimental-setup}

We adopt the same simulation framework as in previous work: a swarm of 4 Toyota HSR robots navigate within a \SI{625}{\meter\squared} closed environment resembling a conference venue, where 6 or 50 people move freely.
The environment and people are simulated in Unity, while robot movements are simulated in ARGoS3~\cite{PinTriOgr-etal2012SI} and mirrored in Unity via ROS communication.
Additional details can be found in previous work~\cite{KegGhaGar-etal2025arxiv}.

Experiments with 8 robots could not be conducted due to resources limitations---primarily memory constraints---but previous results indicate that systems with 4 and 8 robots yield comparable  conclusions.
Each robot is equipped with a forward-facing camera and follows a random walk while continuously performing detection, tracking, and description generation.

We evaluate the performance of the proposed textual descriptions-based swarm perception method (\desc) against that of the previous visual embeddings-based swarm perception baseline (\emb).
With \desc, each robot maintains a local database of text-based clusters, which are evaluated post-experiment using the same metrics as in \emb: cumulative matching characteristic (CMC) curve, mean average precision (mAP), and average cluster purity.
Cluster purity is computed as the proportion of the most represented ground-truth ID within each cluster.
If cases where a ground-truth IDs appears in multiple clusters, only the largest cluster is retained for evaluation. 
Undetected IDs are penalized by assigning them a purity score of 0\%.
This protocol allows for a direct comparison between the language-based and embedding-based approaches, highlighting their respective strengths and limitations.

To assess the impact of communication, we evaluate the 6-person scenario both with and without inter-robot data exchange.
In the 50-person scenario, we test two environmental configurations: one with unobstructed visibility and another featuring large static obstacles that partially occlude each robot’s field of view.

In addition to quantitative evaluation, we conduct qualitative demonstrations of natural-language querying to to assess the interpretability and usability of the system.
To this end, each of the four robots is prompted with a sample description, and returns an image from the cluster that best matches the query.
Results are qualitatively assessed based on the relevance of the retrieved image to the query prompt.

\section{RESULTS AND DISCUSSION}
\label{sec:results}
\begin{figure*}[tbp]
\centering
\includegraphics[width=0.6\textwidth]{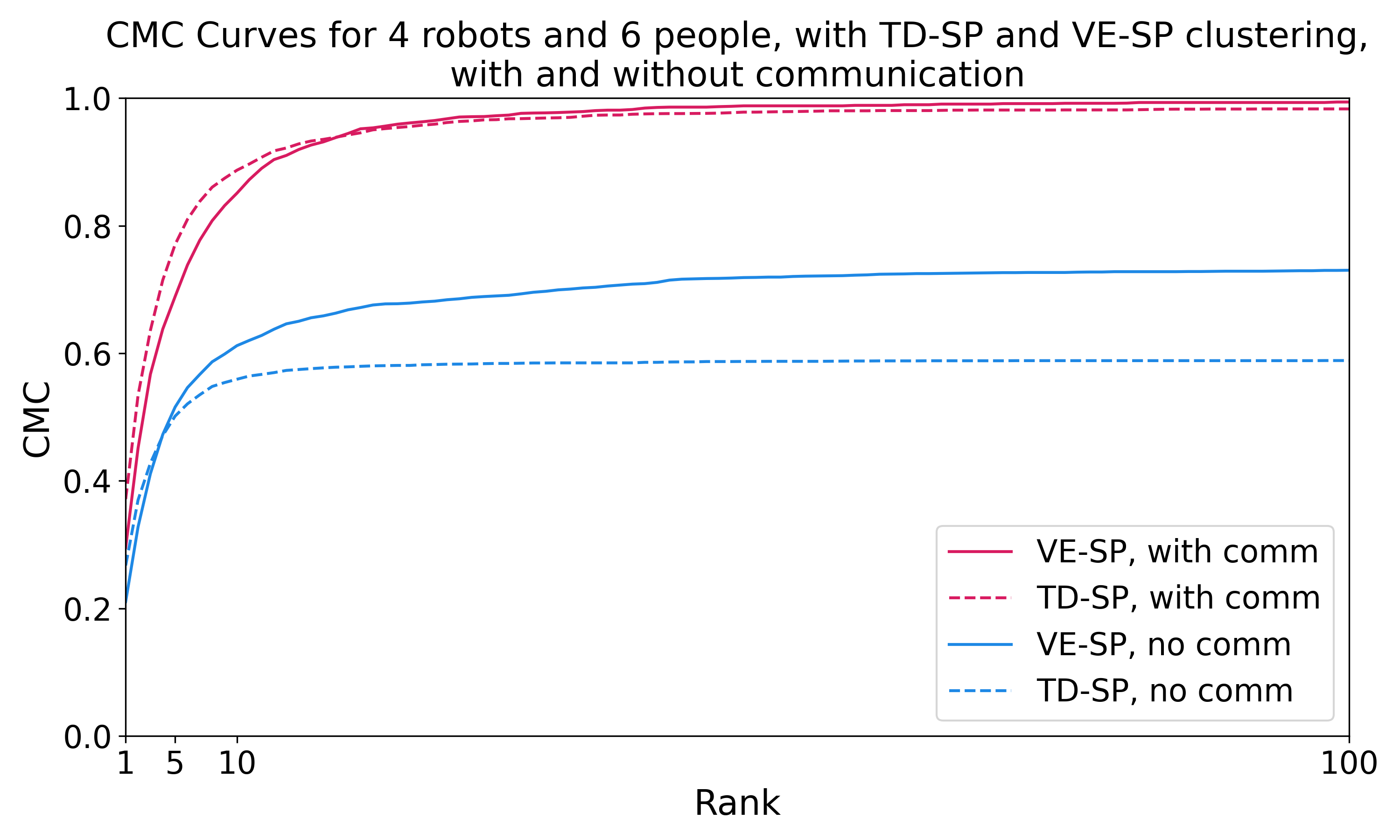}
\hfil
\includegraphics[width=0.35\textwidth]{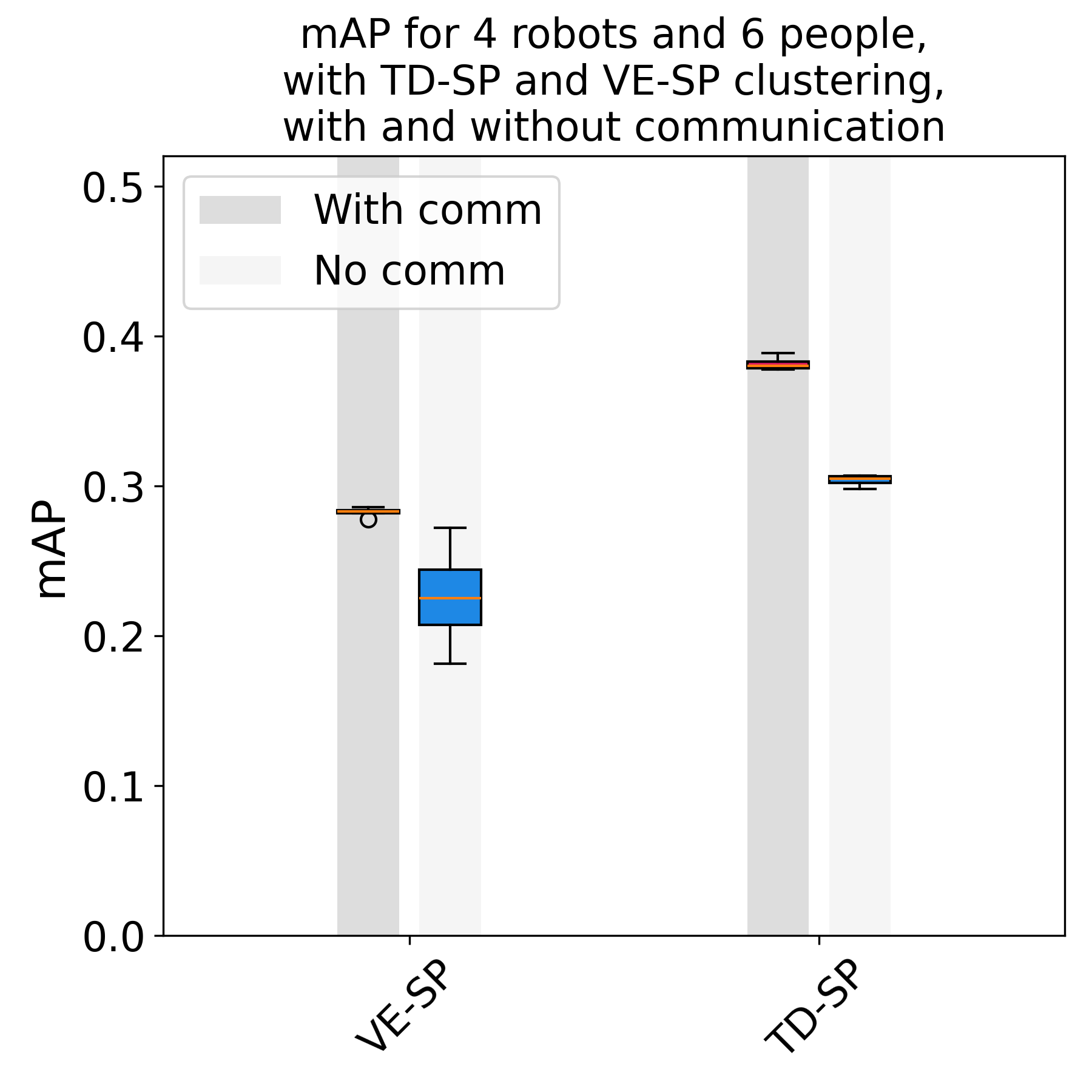}
\hfil
\includegraphics[width=0.4\textwidth]{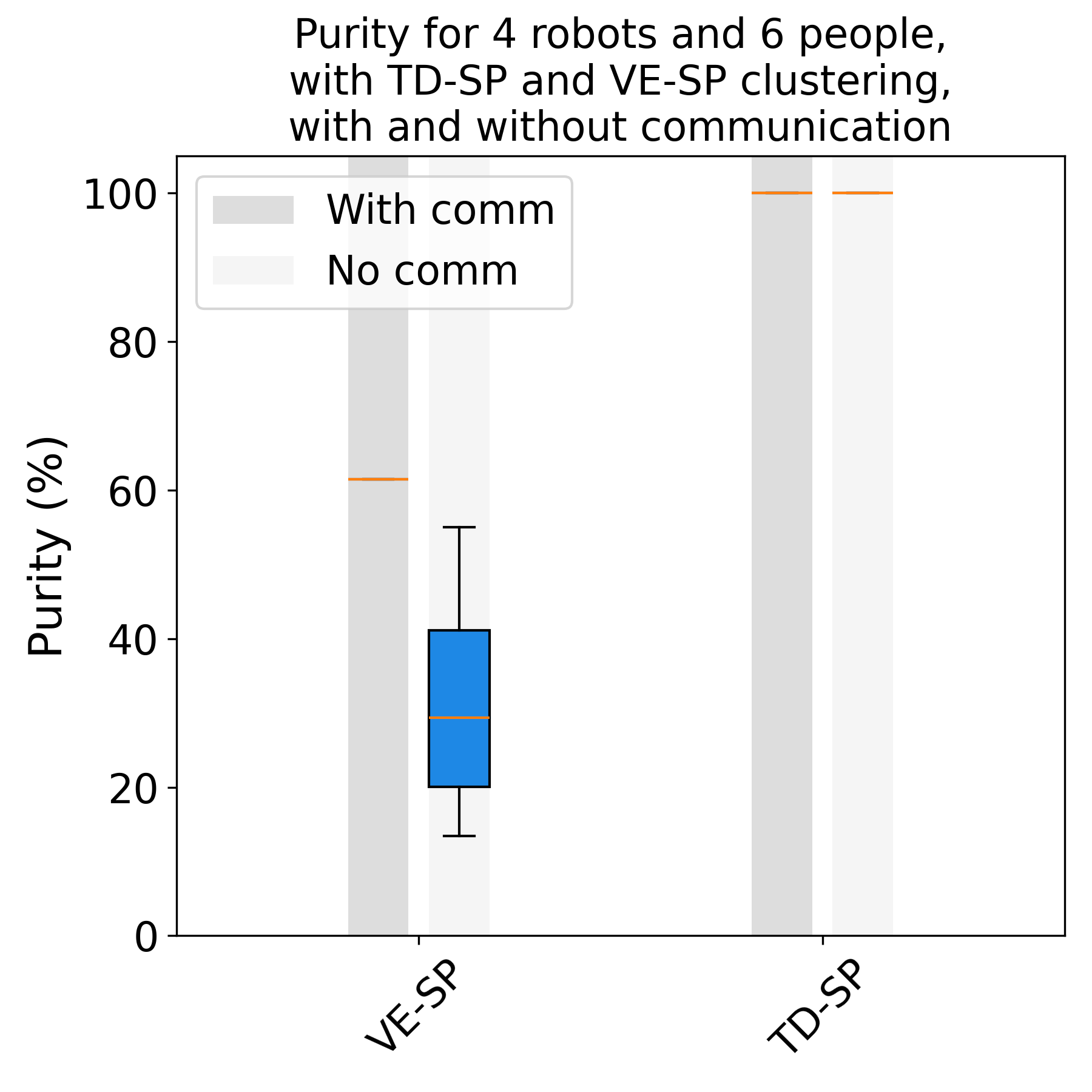}
\hfil
\includegraphics[width=0.4\textwidth]{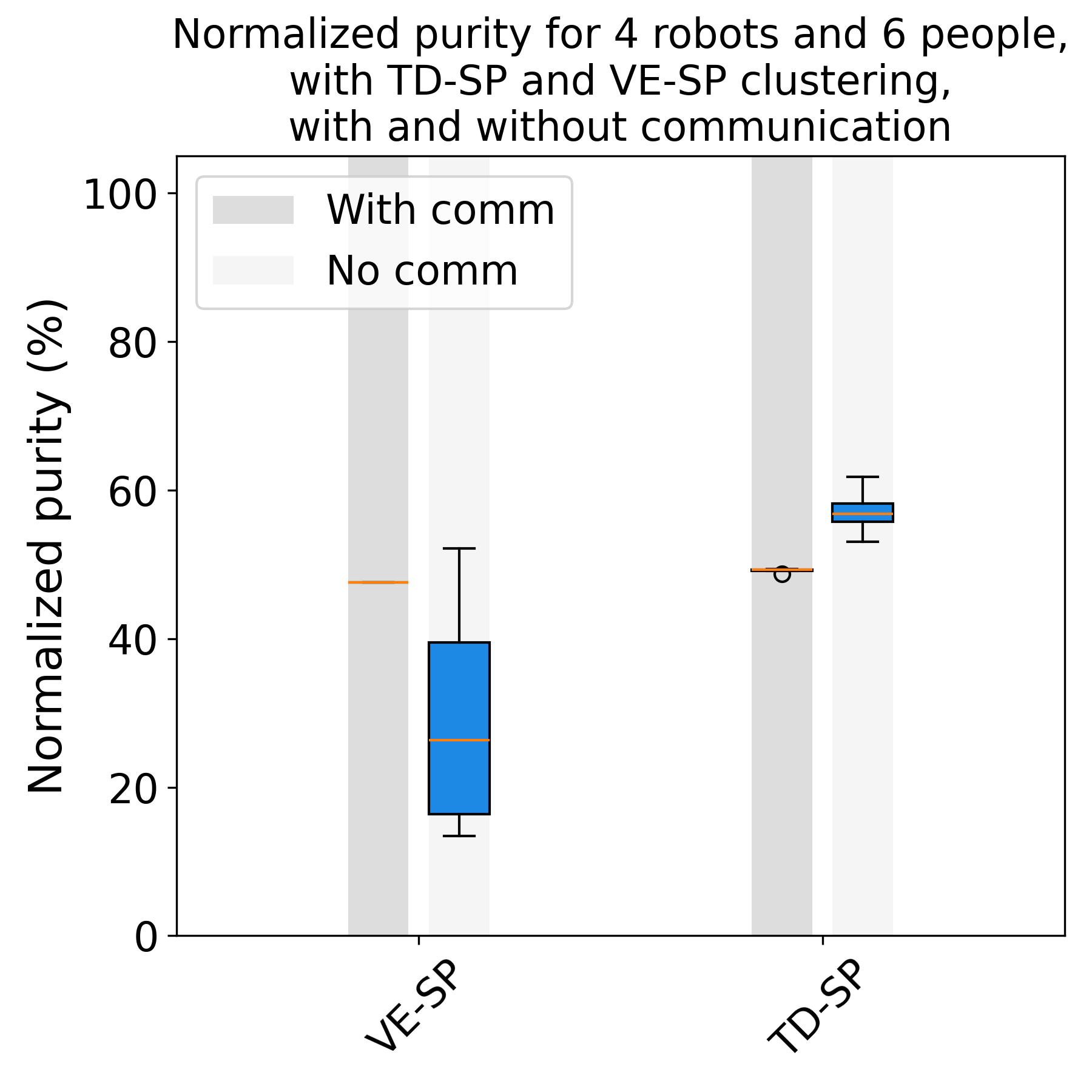}
\hfil
\caption{CMC, mAP, and purity with and without communication for 4 robots and 6 people.}
\label{fig:results_comm}
\end{figure*}
\begin{figure*}[tbp]
\centering
\includegraphics[width=0.6\textwidth]{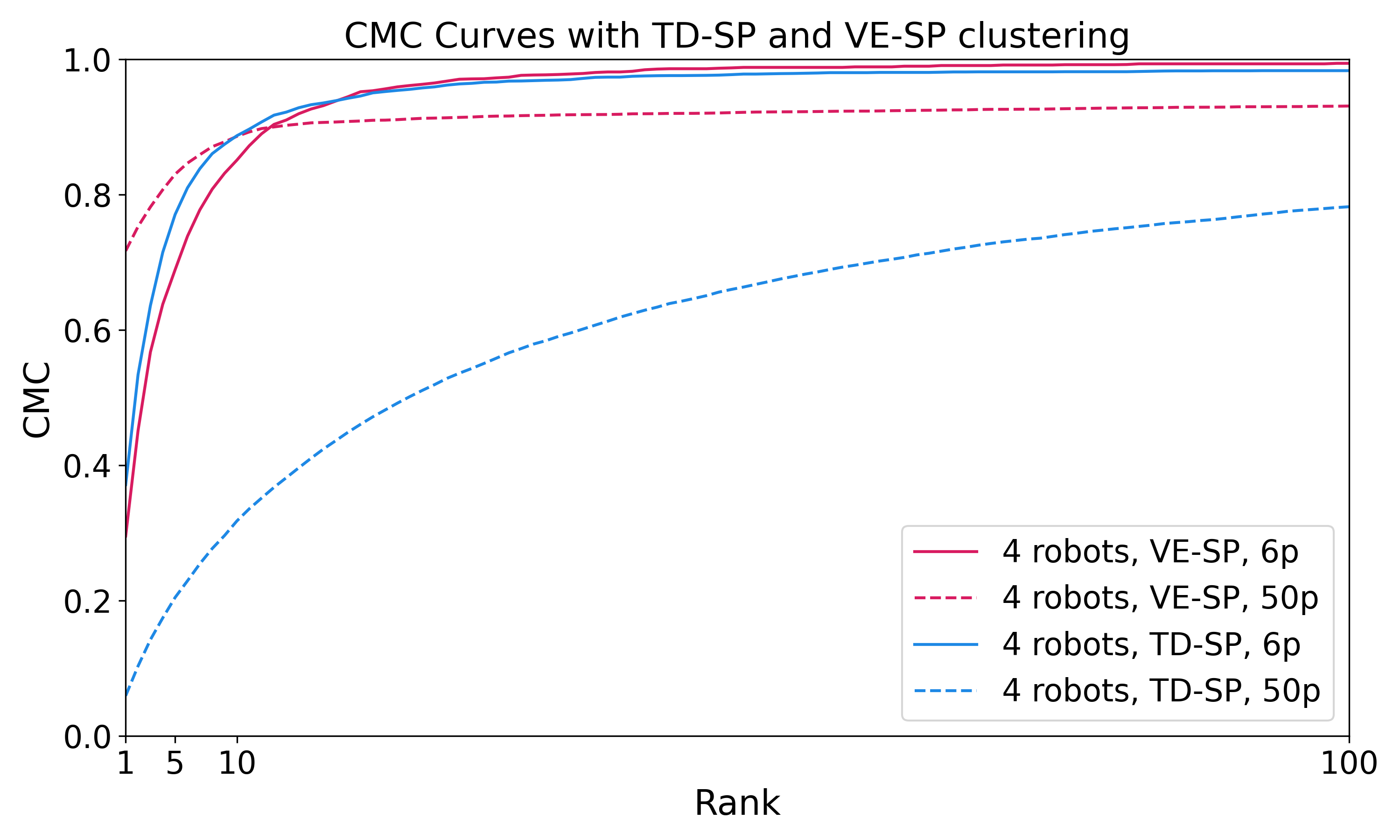}
\hfil
\includegraphics[width=0.35\textwidth]{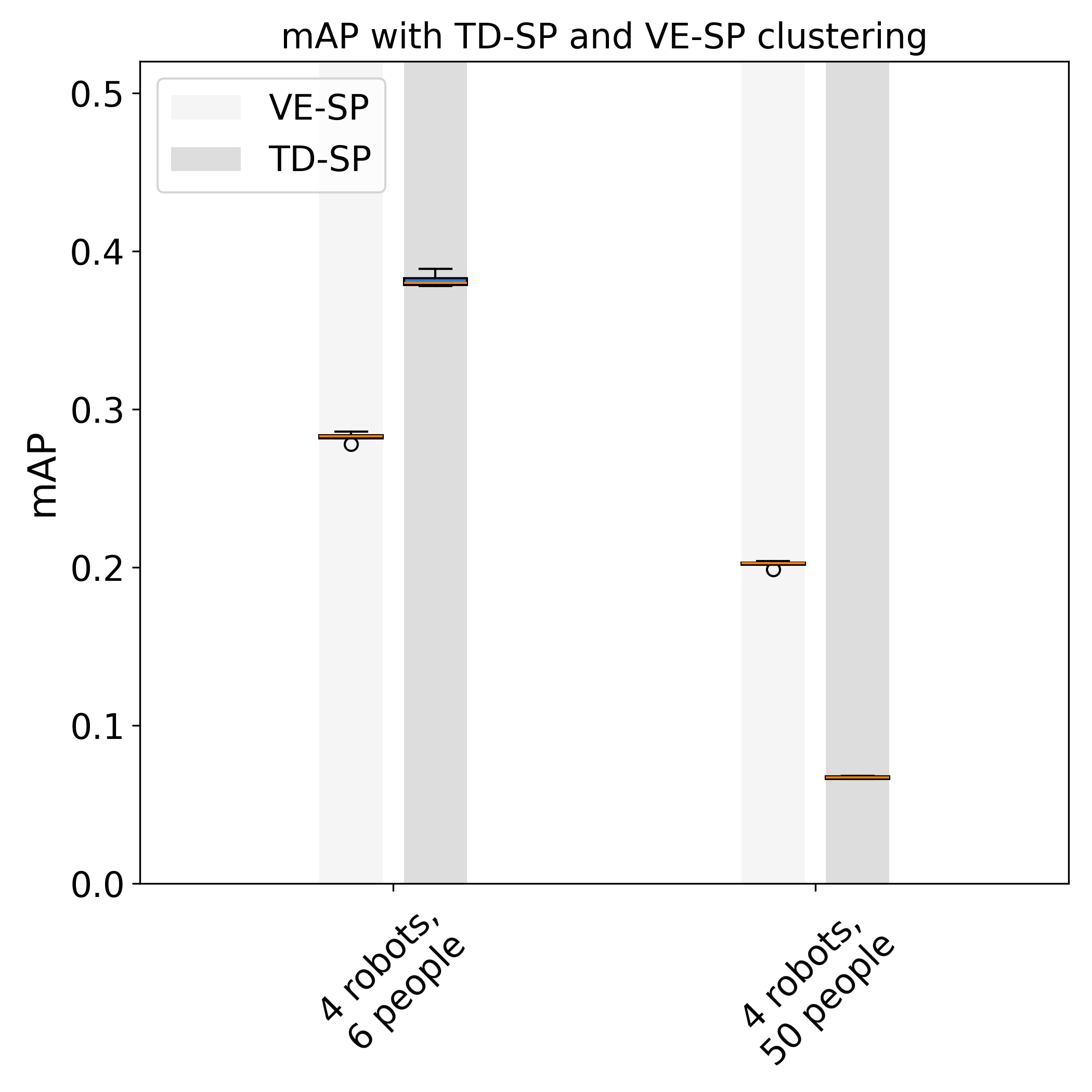}
\hfil
\includegraphics[width=0.4\textwidth]{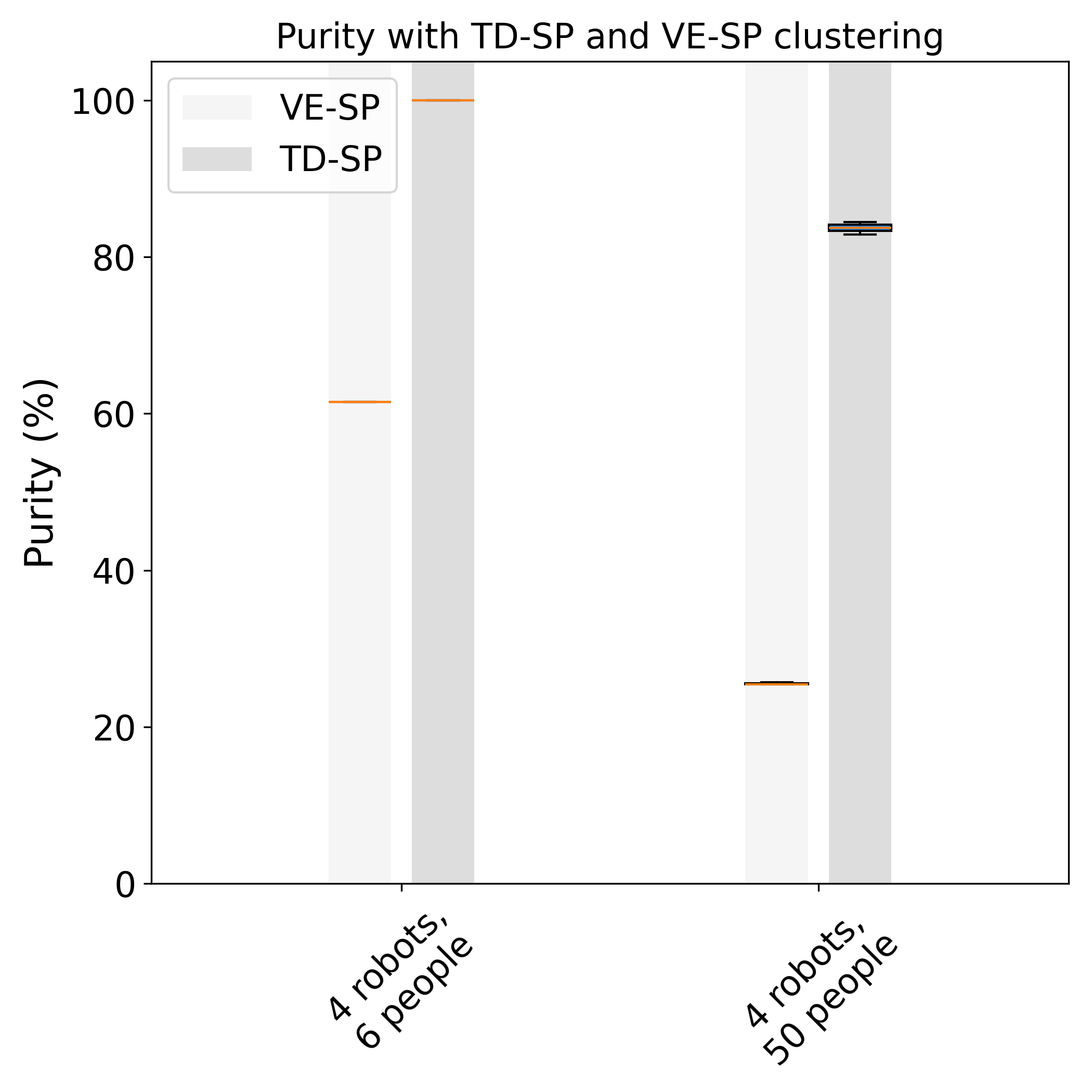}
\hfil
\includegraphics[width=0.4\textwidth]{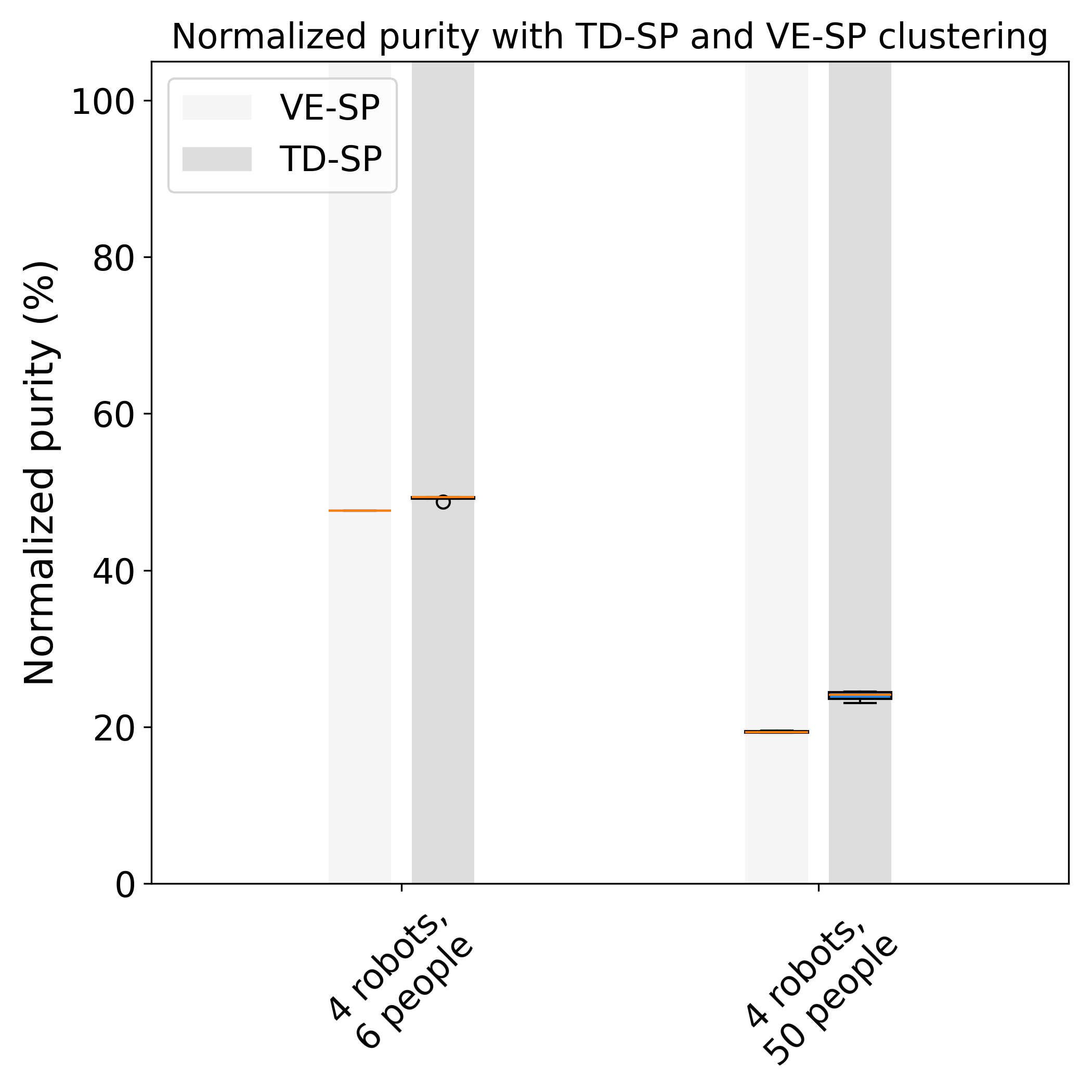}
\hfil
\caption{CMC, mAP, and purity for 4 robots and 6 or 50 people.}
\label{fig:results_people}
\end{figure*}
\begin{figure*}[tbp]
\centering
\includegraphics[width=0.6\textwidth]{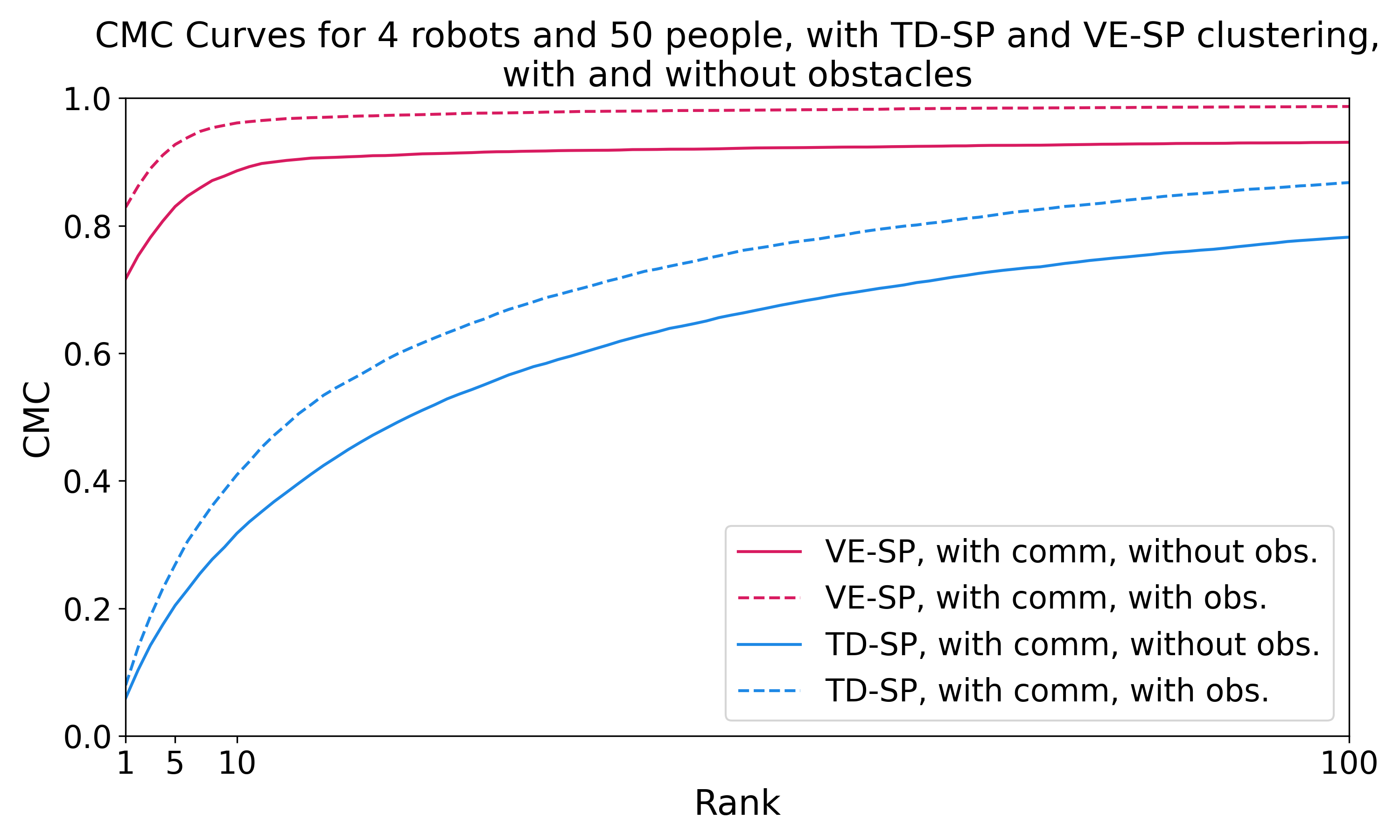}
\hfil
\includegraphics[width=0.35\textwidth]{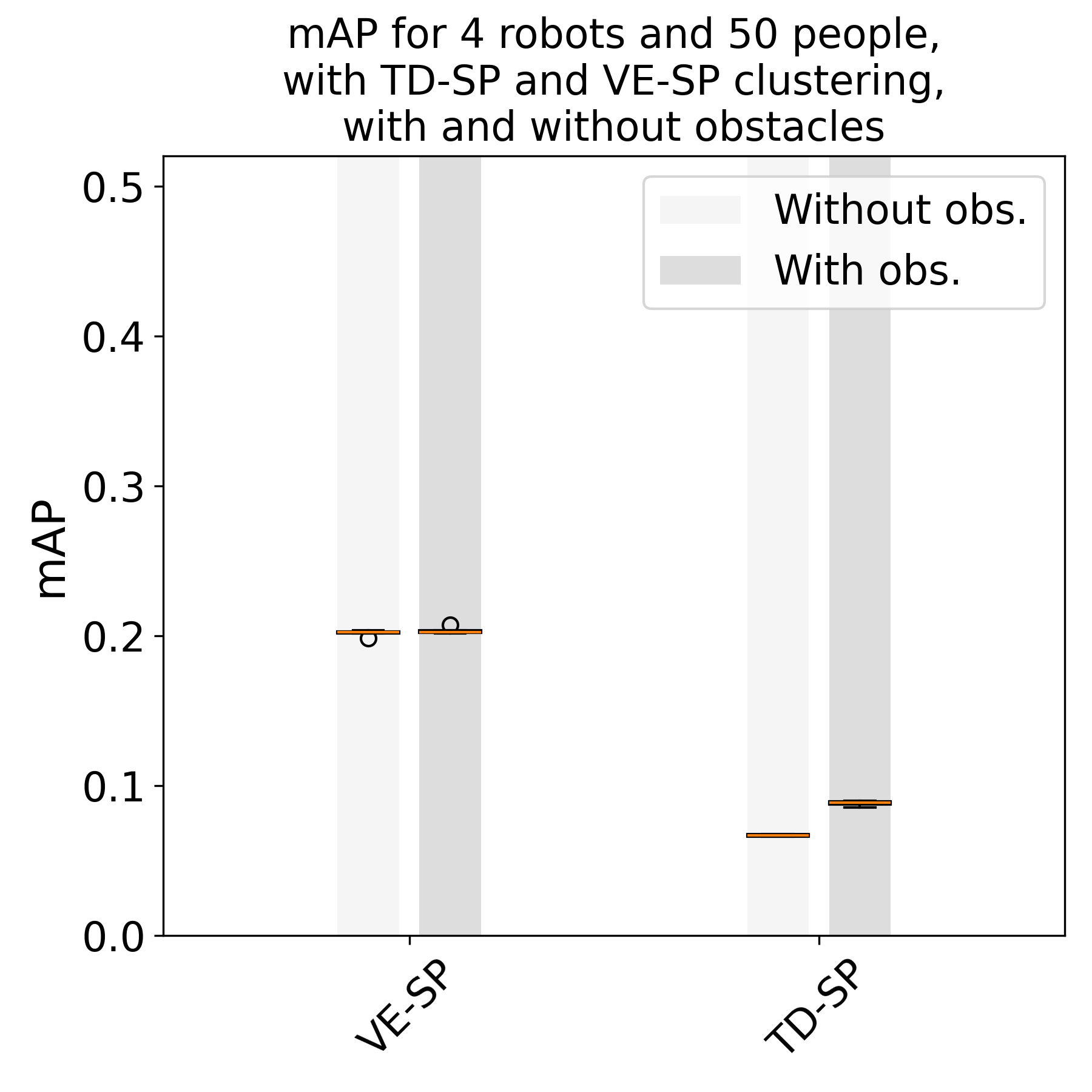}
\hfil
\includegraphics[width=0.4\textwidth]{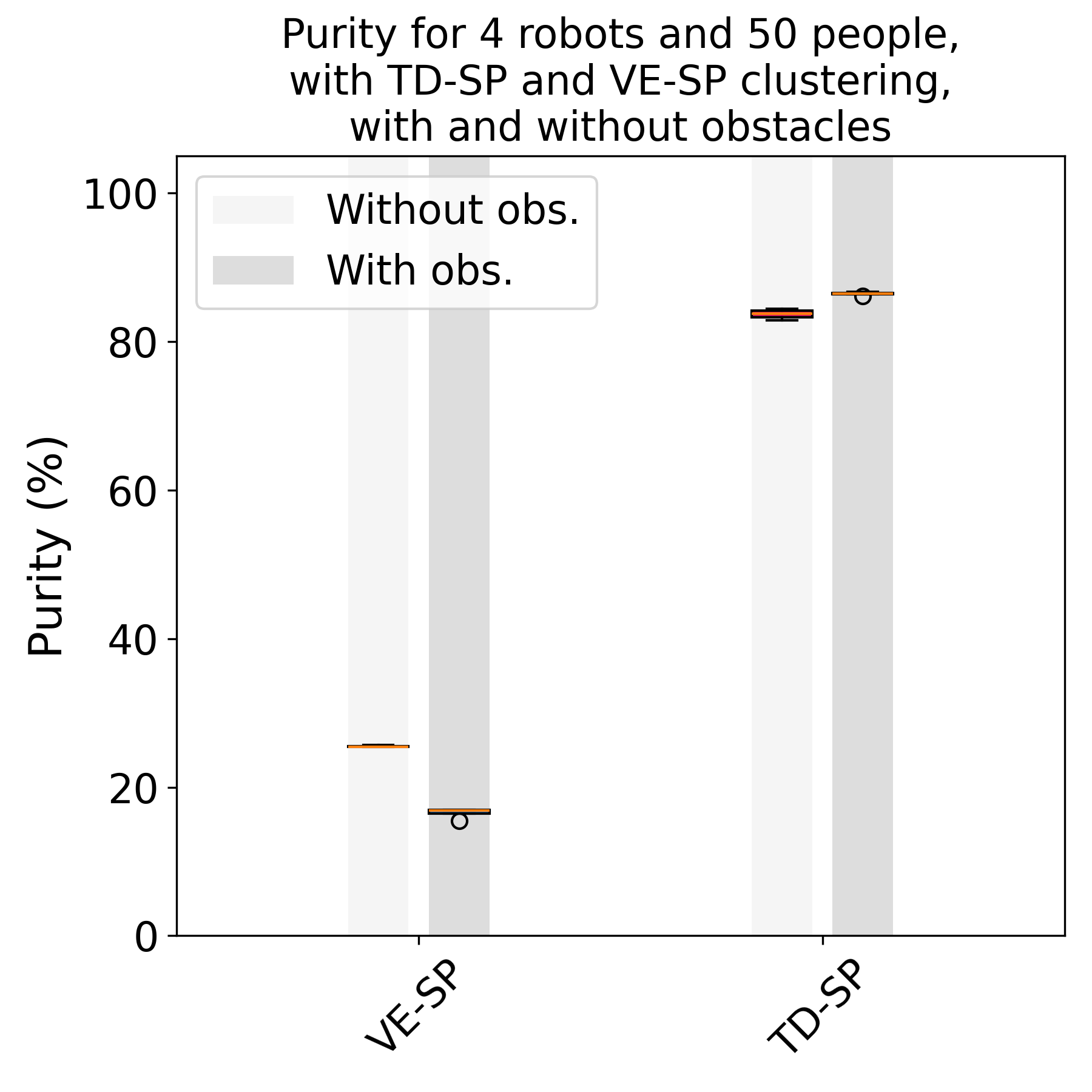}
\hfil
\includegraphics[width=0.4\textwidth]{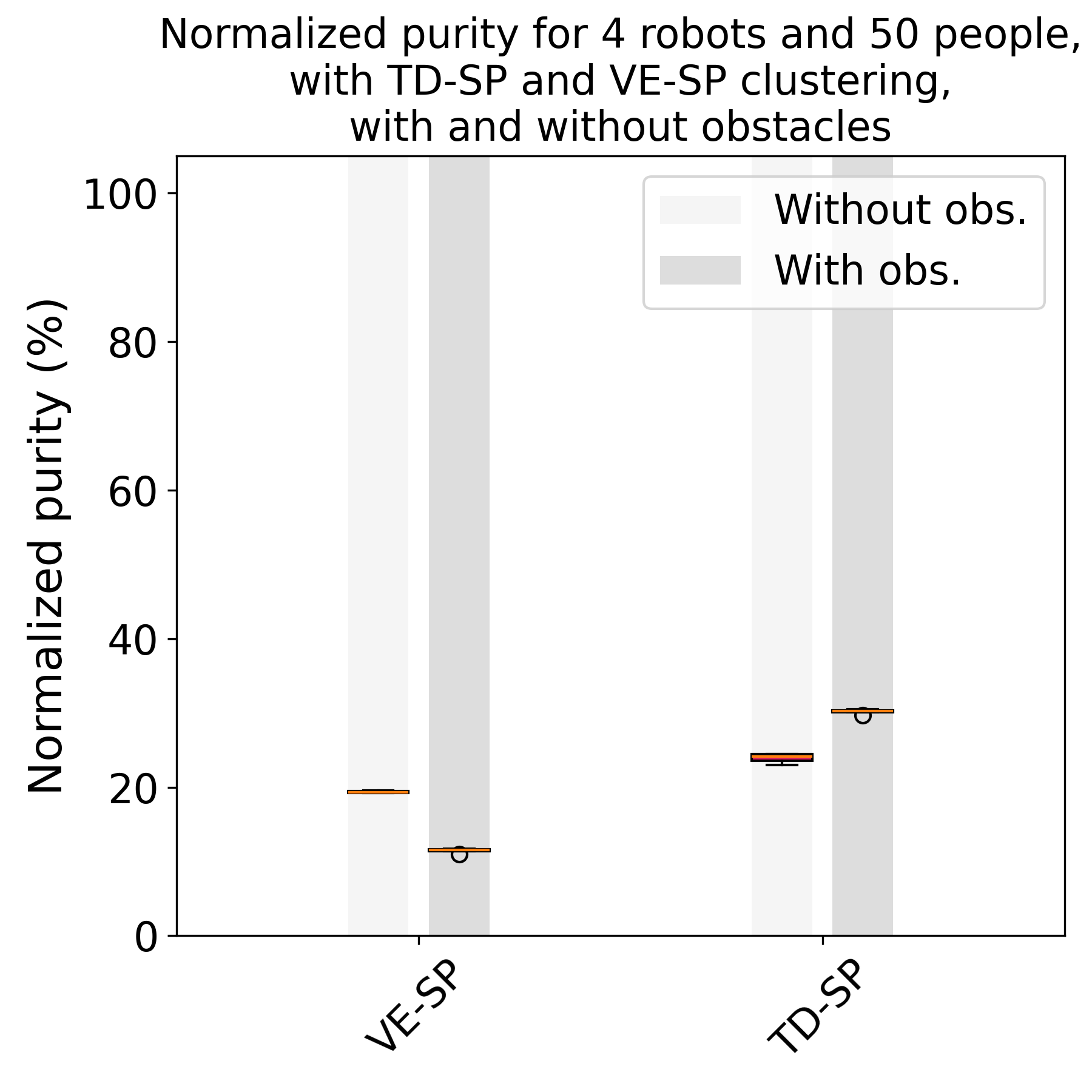}
\hfil
\caption{CMC, mAP, and purity with and without obstacles for 4 robots and 50 people.}
\label{fig:results_obs}
\end{figure*}

The 6-person experiments show that \desc achieves lower or comparable CMC, but higher mAP and significantly higher purity than \emb (see Figure~\ref{fig:results_comm}).
As with \emb, communication between robots improves performance across all metrics.
These results indicate that \desc is a promising approach for swarm-based re-identification, though the lower CMC values point to trade-offs in retrieval ranking performance.

In the 50-person experiments, a more pronounced performance gap emerges.
\desc exhibits a substantial decline in both CMC and mAP, whereas \emb maintains stronger CMC performance and only a modest drop in mAP (see Figure~\ref{fig:results_people}).
Purity decreases for both methods, but the drop is less severe for \desc, which maintains relatively strong results.

To assess robustness to occlusion, we compare performance with and without visual obstacles in the 50-person scenario (see Figure~\ref{fig:results_obs}).
While\emb experiences a drop in purity, \desc remains largely unaffected by obstacles and even shows modest improvements across all metrics. This may be attributed to the language-based model's capacity to generalize from partially visible features, or to the increased viewpoint diversity introduced by constrained navigation paths.

Overall, these results suggest that \desc is a viable alternative to \emb, offering better interpretability and higher cluster purity, albeit with limitations that affect scalability and ranking-based metrics.
One key issue is over-fragmentation:
\desc tends to produce significantly more clusters than \emb, likely due to the rigid nature of cosine similarity over sentence embeddings.
This leads to many fine-grained clusters that individually exhibit high purity but collectively reduce overall matching accuracy.

To further investigate this, we computed normalized cluster purity, which averages the purity of all clusters associated with each ground-truth ID, rather than only considering the largest.
While this adjustment yields only a slight reduction for \emb, the score for \desc drops more noticeably---though it still outperforms \emb.
This confirms that \desc creates many clusters that are locally accurate but globally redundant, leading to lower CMC and mAP.
As the number of individuals increases, this fragmentation becomes more severe, resulting in degraded downstream performance.
We identify this over-clustering as a key bottleneck for future improvement.

Replacing cosine similarity with more flexible matching mechanisms is a natural direction forward.
Preliminary tests with cross-encoders show promise in reducing fragmentation but also introduce significant computational overhead, making them impractical for large-scale experiments at present.

\subsection*{Natural Language Query Evaluation}

We qualitatively assess the ability of users to retrieve information from the swarm using natural-language prompts
In both the 6- and 50-person scenarios (with communication and no obstacles), each robot is queried with three example descriptions:
\begin{itemize}
    \item Query 1: a lady with a green t-shirt;
    \item Query 2: a person with red shirt and black skirt;
    \item Query 3: a person with a black outfit.
\end{itemize}
Each robot returns the image from its local cluster that best matches the query.
Note that although our method operates entirely on textual descriptions, we enabled the sharing of cropped person images between robots specifically for this demonstration. 
This was done solely to allow robots to return images received from their peers---enabling a more comprehensive qualitative assessment---and does not influence the functioning of the method itself.

\begin{figure*}[tbp]
\centering

\begin{minipage}[b]{\textwidth}
\centering
\small Query~1: a lady with a green t-shirt
\\[1.5mm]
\includegraphics[height=0.18\textwidth]{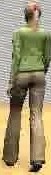}
\hfil
\includegraphics[height=0.18\textwidth]{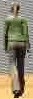}
\hfil
\includegraphics[height=0.18\textwidth]{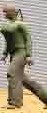}
\hfil
\includegraphics[height=0.18\textwidth]{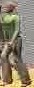}
\end{minipage}
\\[8mm]

\begin{minipage}[b]{\textwidth}
\centering
\small Query~2: a person with red shirt and black skirt
\\[1.5mm]
\includegraphics[height=0.18\textwidth]{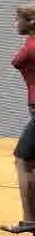}
\hfil
\includegraphics[height=0.18\textwidth]{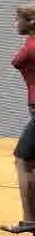}
\hfil
\includegraphics[height=0.18\textwidth]{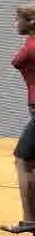}
\hfil
\includegraphics[height=0.18\textwidth]{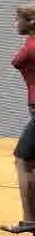}
\end{minipage}
\\[8mm]

\begin{minipage}[b]{\textwidth}
\centering
\small Query~3: a person with a black outfit
\\[1.5mm]
\includegraphics[height=0.18\textwidth]{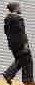}
\hfil
\includegraphics[height=0.18\textwidth]{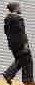}
\hfil
\includegraphics[height=0.18\textwidth]{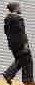}
\hfil
\includegraphics[height=0.18\textwidth]{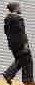}
\end{minipage}

\caption{Outputs for 6-people experiment from Robot 0 to Robot 3 (from left to right) for different queries.}
\label{fig:results_queries_6}
\end{figure*}

\begin{figure*}[tbp]
\centering

\begin{minipage}[b]{\textwidth}
\centering
\small Query~1: a lady with a green t-shirt
\\[1.5mm]
\includegraphics[height=0.18\textwidth]{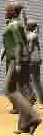}
\hfil
\includegraphics[height=0.18\textwidth]{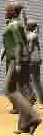}
\hfil
\includegraphics[height=0.18\textwidth]{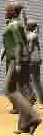}
\hfil
\includegraphics[height=0.18\textwidth]{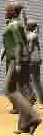}
\end{minipage}
\\[8mm]

\begin{minipage}[b]{\textwidth}
\centering
\small Query~2: a person with red shirt and black 
\\[1.5mm]
\includegraphics[height=0.18\textwidth]{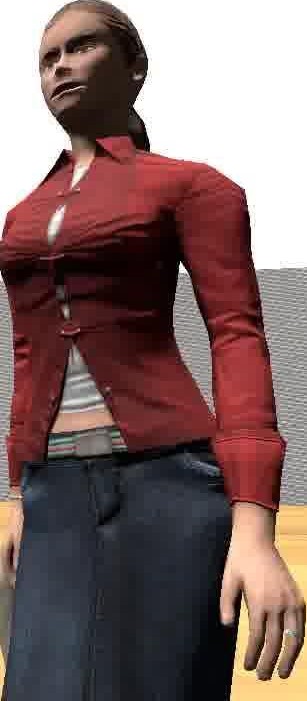}
\hfil
\includegraphics[height=0.18\textwidth]{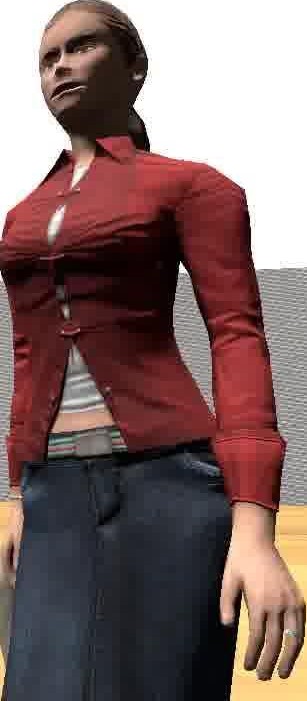}
\hfil
\includegraphics[height=0.18\textwidth]{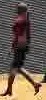}
\hfil
\includegraphics[height=0.18\textwidth]{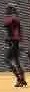}
\end{minipage}
\\[8mm]

\begin{minipage}[b]{\textwidth}
\centering
\small Query~3: a person with a black outfit
\\[1.5mm]
\includegraphics[height=0.18\textwidth]{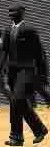}
\hfil
\includegraphics[height=0.18\textwidth]{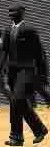}
\hfil
\includegraphics[height=0.18\textwidth]{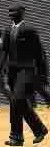}
\hfil
\includegraphics[height=0.18\textwidth]{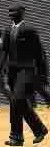}
\end{minipage}

\caption{Outputs for 50-people experiment from Robot 0 to Robot 3 (from left to right) for different queries.}
\label{fig:results_queries_50}
\end{figure*}

As shown in Figures~\ref{fig:results_queries_6} and \ref{fig:results_queries_50}, all robots consistently retrieve images of the correct individual.
Interestingly, for Query~1 (6 people) and Query~2 (50 people), robots return different images, suggesting that each had independently formed a reliable cluster before communication occurred.
In contrast, for the remaining queries, the robots return identical images---indicating that the underlying cluster likely originated from a single robot and was propagated to others via communication.
Specifically, Robot~0 was the likely origin for Query~2 (6 people), Robot~2 for Query~3 (6 people), Robot~1 for Query~1 (50 people), and Robot~2 for Query~3 (50 people).
This supports the conclusion that the system enables both independent perception and effective sharing of semantic information across the swarm.

%
\section{CONCLUSIONS AND FUTURE WORK}
\label{sec:conclusions}

In this work, we introduced a decentralized person re-identification method based on natural language descriptions in place of opaque visual feature embeddings. 
This approach enhances swarm perception by making internal representations more interpretable and enabling intuitive, language-based querying. 
While current performance is constrained by processing limitations and rigid clustering thresholds, results demonstrate the viability of language-driven re-identification in decentralized systems.

Future work will focus on addressing current limitations and expanding the system's capabilities:
\begin{itemize}
    \item Developing more flexible similarity measures, such as learned matching functions or transformer-based clustering.
    \item Exploring lightweight language models suitable for real-time, onboard deployment.
    \item Extending perception to include environmental context (e.g., “a person in red was seen near a table”).
    \item  Investigating selective communication strategies to reduce bandwidth and memory overhead.
    \item Integrating visual and language-based modalities to improve robustness and disambiguation.
\end{itemize}
The proposed method opens new directions for interpretable, scalable, and human-accessible swarm perception, supporting more natural interaction between humans and decentralized robotic systems.

\bibliographystyle{IEEEtran}
\bibliography{demiurge-bib/definitions,demiurge-bib/author,demiurge-bib/address,demiurge-bib/proceedings-short,demiurge-bib/journal-short,demiurge-bib/publisher,demiurge-bib/series-short,demiurge-bib/institution,demiurge-bib/bibliography,additions}

\end{document}